\pdfoutput=1
\documentclass[11pt]{article}
\usepackage[table]{xcolor}

\usepackage{acl}

\usepackage{times}
\usepackage{latexsym}

\usepackage[T1]{fontenc}
\usepackage[utf8]{inputenc}
\usepackage{graphicx}
\usepackage{dsfont}
\usepackage{amssymb}

\usepackage{microtype}

\usepackage{booktabs} % For better table formatting
\usepackage{multirow} % For multi-row cells
\usepackage{graphicx} % For rotating text (e.g., for the header of the rows)
\usepackage{amsmath}
\newcommand{\tabincell}[2]{\begin{tabular}{@{}#1@{}}#2\end{tabular}}
\makeatletter
\renewcommand{\maketag@@@}[1]{\hbox{\m@th\normalsize\normalfont#1}}%
\makeatother

\title{VideoCLIP-XL: Advancing Long Description Understanding for Video CLIP Models}

\author{Jiapeng Wang$^{1}$\thanks{\ \ Contribution during internship at Alibaba Cloud Computing.}, Chengyu Wang$^{2}$\thanks{\ \ Co-corresponding authors.}, Kunzhe Huang$^{2}$, Jun Huang$^{2}$, Lianwen Jin$^{1}$\footnotemark[2]\\
  $^{1}$South China University of Technology, Guangzhou, China \\
  $^{2}$Alibaba Cloud Computing, Hangzhou, China \\
  \texttt{eejpwang@mail.scut.edu.cn, eelwjin@scut.edu.cn}\\
  \texttt{\{chengyu.wcy, huangkunzhe.hkz, huangjun.hj\}@alibaba-inc.com}
}

\begin{document}
\maketitle
\begin{abstract}
Contrastive Language-Image Pre-training (CLIP) has been widely studied and applied in numerous applications.
However, 
the emphasis on brief summary texts during pre-training prevents CLIP from understanding long descriptions.
This issue is particularly acute regarding videos 
given that videos often contain abundant detailed contents.
In this paper, we propose the \textbf{VideoCLIP-XL}  (\textit{eXtra Length}) model, which aims to unleash the long-description understanding capability of video CLIP models.
Firstly, we establish an automatic data collection system and gather a large-scale \textbf{VILD} pre-training dataset\footnote{
\href{https://huggingface.co/alibaba-pai/VILD}{https://huggingface.co/alibaba-pai/VILD}.
} with \textit{VIdeo} and \textit{Long-Description} pairs. 
Then, we propose Text-similarity-guided Primary Component Matching (\textbf{TPCM}) to better learn the distribution of feature space while expanding the long description capability.
We also introduce two new tasks namely Detail-aware Description Ranking (\textbf{DDR}) and Hallucination-aware Description Ranking (\textbf{HDR}) for further understanding improvement.
Finally, we construct a Long Video Description Ranking (\textbf{LVDR}) benchmark\footnote{
\href{https://huggingface.co/alibaba-pai/LVDR}{https://huggingface.co/alibaba-pai/LVDR}.
} for evaluating the long-description capability more comprehensively.
Extensive experimental results on widely-used text-video retrieval benchmarks with both short and long descriptions and our LVDR benchmark can fully demonstrate the effectiveness of our method.\footnote{
\href{https://huggingface.co/alibaba-pai/VideoCLIP-XL}{https://huggingface.co/alibaba-pai/VideoCLIP-XL}.
}

\end{abstract}

\section{Introduction}
The Contrastive Language-Image Pre-training (CLIP) model~\cite{clip} represents a pivotal development in the field of vision-language pre-training. It integrates text and image encoders to align these two modalities through contrastive learning. This methodology has been effectively applied in various applications, such as zero-shot classification~\cite{alphaclip}, text-image retrieval~\cite{lex}, and text-to-image generation~\cite{stablediffusion, clipdraw}. However, one of the notable limitations of CLIP is its constrained capacity to process extensive textual descriptions, owing to its text encoder's reliance on maximum positional embeddings with length 77. 
This limitation  greatly
restricts the length of input text, and
existing studies~\cite{longclip} have also revealed a \emph{de facto} effective token limit of just around 20.

Furthermore, the vanilla CLIP training procedure's emphasis on brief summary texts compels the text/vision encoder to focus predominantly on the main features of the text/visual input, often overlooking smaller, yet potentially critical details. 
This issue is \textit{particularly acute in videos} as compared to images, given that videos encapsulate a wealth of details across successive frames, along with additional information such as the sequence and flow of activities, camera movements, etc. In this context, existing video CLIP models~\cite{videoclip,clip4clip,viclip} that employ vanilla CLIP training methodologies may struggle to accurately capture complex relationships and attributes, due to their reliance on a simplistic ``bag of concepts'' approach~\cite{lemon}.
To overcome these limitations, enhancing the model's capability to comprehend long descriptions is crucial. Longer texts provide a rich tapestry of attributes and interconnections, offering a pathway to significantly improve the model's performance and applicability in more complex scenarios.

To this end, we propose \textbf{VideoCLIP-XL} (\textit{eXtra Length}), to our knowledge, which is the first video CLIP model with long-description capability. 
(1)
To be specific, recognizing the insufficiency of public datasets containing (\textit{video}, \textit{long description}) pairs, we establish an automatic data collection system, designed to aggregate sufficient and high-quality pairs from multiple data sources. We have successfully collected over 2M (\textit{\textbf{VI}deo}, \textit{\textbf{L}ong \textbf{D}escription}) pairs, denoted as our \textbf{VILD} pre-training dataset.
(2) 
We have discovered that existing CLIP models for long texts~\cite{longclip} lack the flexibility to dynamically adapt to the distribution changes within high-dimensional feature space.
To address this issue, we introduce \textbf{T}ext-similarity-guided \textbf{P}rimary \textbf{C}omponent \textbf{M}atching (\textbf{TPCM}), a novel approach that enables the model to better learn cross-modal  and  cross-sample relative distances. 
(3) We claim that there are two attributes that CLIP models with the long-description understanding capability should naturally possess: for a given video and its associated descriptions, it should be able to assign a higher score when the description contains i) more rich and precise detailed contexts; or ii) fewer hallucinations with the same level of detail.
To this end, we propose two new tasks to model these two attributes namely \textbf{D}etail-aware \textbf{D}escription \textbf{R}anking (\textbf{DDR}) and \textbf{H}allucination-aware \textbf{D}escription \textbf{R}anking (\textbf{HDR}).
They make the video CLIP model to learn how to correctly rank multiple descriptions with different levels of details and hallucinations.
(4) In order to better evaluate video CLIP models,
we further release a \textbf{L}ong \textbf{V}ideo \textbf{D}escription \textbf{R}anking (\textbf{LVDR}) 
benchmark.
Given each video and the corresponding ground-truth long description (after human correction) sampled from Shot2Story~\cite{shot2story},
we iteratively modify a certain proportion of correct contents into hallucination in each step.
The model is required to correctly rank these descriptions according to their faithfulness.

To evaluate the performance of VideoCLIP-XL, we conduct extensive experiments not only on the video \& long-description dataset Shot2Story~\cite{shot2story}, but also on traditional widely-used 
MSR-VTT~\cite{xu2016msr},  LSMDC~\cite{lsmdc},  DiDeMo~\cite{didemo},  MSVD~\cite{msvd} and  ActivityNet~\cite{caba2015activitynet} 
benchmarks, for the text-video retrieval task.
Moreover, we evaluate VideoCLIP-XL and other representative CLIP models on our proposed LVDR benchmark.
Experimental results demonstrate that our method exhibits superior performance compared with state-of-the-art competitors. 

Our main contributions are as follows:
\begin{itemize}
\item We propose the VideoCLIP-XL model to unleash the long-description understanding capability of video CLIP models. We also collect and release a new pre-training dataset VILD with over 2M video \& long-description pairs using our automatic data collection system.
\item In VideoCLIP-XL, we propose TPCM 
for dynamic feature learning
while expanding the long description capability.
We also propose two new tasks (i.e., DDR and HDR) to further model the effective attributes for better representation learning of long descriptions.
\item To better evaluate video CLIP models' long description ability, 
we propose the LVDR benchmark for long description ranking. 
\item Extensive experiments show that VideoCLIP-XL clearly outperforms state-of-the-art models over various tasks and benchmarks.
\end{itemize}

\section{Related Work}
\noindent\textbf{Image/Video CLIP models.}
CLIP~\cite{clip} is a multimodal model based on contrastive learning. Its training data comprises a vast collection of text-image pairs, each image paired with a corresponding text description. Through contrastive learning, the model learns the matching relationship between text-image pairs. Owing to its robust zero-shot generalization capabilities, CLIP has been successfully deployed in numerous scenarios including detection~\cite{vild, glip}, segmentation~\cite{groupvit, lseg}, image/video understanding~\cite{clip4clip, videoclip, xtremeclip}, retrieval~\cite{cocaclip,fashionklip} and image generation~\cite{DALLE2, clipdraw, vqganCLIP, CLIPpasso}. 
For video analysis, ViCLIP~\cite{viclip} incorporates spatio-temporal attention within its video encoder and adopts partial random patch masking during training. Nonetheless, several subsequent studies~\cite{rovit, xvlm} have identified CLIP's inadequacy in extracting fine-grained information. These works implement contrastive methods similar to CLIP's to align complete sentence tokens with regions of the entire image.
Furthermore, Long-CLIP~\cite{longclip} proposes the use of primary component matching of CLIP features to improve the model's understanding of lengthy descriptions in images.

\noindent\textbf{Vision-Language Datasets.} 
As the capabilities of multimodal models advance, the need transcends traditional fixed-category image datasets such as ImageNet~\cite{image} and CIFAR10~\cite{cifar}. Contemporary open-world applications require datasets that encompass both images/videos and their associated text descriptions. Common open-world image-language datasets include Visual Genome~\cite{VG}, Conceptual-12M~\cite{cc12m}, SBU~\cite{sbu}, COCO~\cite{COCO}, and LAION-5B~\cite{laion}. Typical video-language datasets comprise MSR-VTT~\cite{xu2016msr}, MSVD~\cite{msvd}, LSMDC~\cite{lsmdc}, WebVid~\cite{webvid}, InternVid~\cite{viclip}, and Panda-70M~\cite{chen2024panda}. However, these datasets generally contain only short captions.
On the other hand, a few datasets focus on long descriptions. ShareGPT4V~\cite{sharegpt4v} is a large-scale dataset with 1.2M long captions for images. Shot2Story~\cite{shot2story} includes 20K video clips, each with detailed shot-level captions and comprehensive video summaries. MiraData~\cite{miradata} deals with uncut video segments and features structural long captions. It contains 57.8K video clips across two scenarios: gaming and city/scenic exploration. 
The average description length in these collections is often orders of magnitude longer than those in previous datasets~\cite{longclip}.

\section{Methodology}
In this section, we introduce our automatic data collection system and the resulting   \textit{VIdeo} \& \textit{Long-Description} (VILD) pre-training dataset (Sect.~\ref{vld}),
the text-similarity-guided primary component matching (TPCM) technique (Sect.~\ref{dpcm}),
two new description ranking tasks (Sect.~\ref{drt}), and the new Long Video Description Ranking (LVDR)
benchmark dataset (Sect.~\ref{lvdr}).

\begin{figure}[t]
\centering
\includegraphics[width=0.49\textwidth]{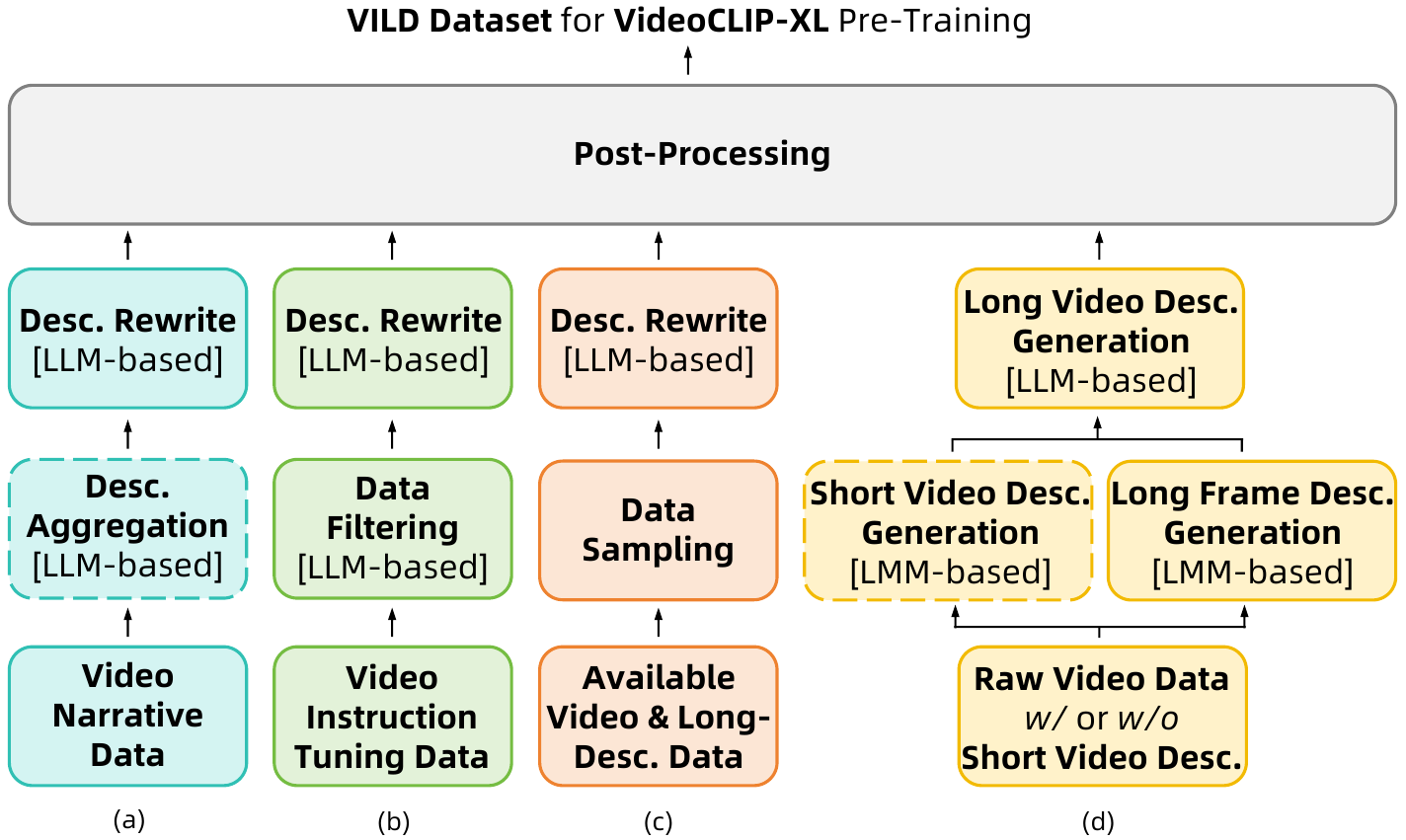}
\caption{The automatic data collection system for our VILD dataset. Desc. is short for description. 
} 
\label{image-vld}
\end{figure}

\subsection{VIdeo \& Long-Description (VILD) Dataset}\label{vld}
Training CLIP models often necessitates a substantial corpus of vision-text pairs. 
In image processing, the advent of open-source large multimodal models (LMMs) and the availability of APIs such as GPT-4V~\cite{GPT4v} have spurred efforts to annotate images with detailed long descriptions. 
For example, ShareGPT4V~\cite{sharegpt4v} is a large dataset, originating from a high-quality curated set of 100K captions gathered using GPT-4V and expanded to 1.2M through a caption model.

However, video datasets with extensive long descriptions, especially in the open domain, remain markedly scarce. 
For instance, Shot2Story~\cite{shot2story} offers 20K video clips, each accompanied by shot-level captions and video summaries. After annotating with LMMs, further manual corrections ensure the reliability of these long descriptions, \textit{thereby qualifying it as a trustworthy evaluation set, which is excluded from our training data}.
MiraData~\cite{miradata} leverages GPT4V to produce long captions for 57.8K video clips, restricted to gaming and city/scenic exploration scenarios.
Open-Sora-Dataset~\cite{opensoradata} utilizes LMMs to generate descriptive narratives for 40.2K videos, predominantly in natural landscapes.

In light of the scarcity of open-domain video \& long-description pairs, we engineer an automatic data collection system, as depicted in Fig.~\ref{image-vld}. 
Our approach harnesses multiple sources, chiefly encompassing video narrative data, video instruction tuning data, raw videos, and available video \& long-description pairs.

\noindent\textbf{(a) Video Narrative Data.} 
Video narrative data often contains human-concerned descriptions produced
by human annotators, which can describe the whole scene, the main activities, and events involving multiple actors and objects.
We adopt the VidLN~\cite{VidLN} dataset, consisting of human-annotated individual-level descriptions of each single main people/animal/objective in the video along with the background.
To make the dataset serve our purpose, we employ large language models (LLMs) to aggregate individual-level narratives into whole-level descriptions through prompt engineering (i.e., the \textit{Desc. Aggregation} step). 
Finally, in consideration of training efficacy and robustness, we further utilize LLMs to rewrite the whole-level description  (i.e., the \textit{Desc. Rewrite} step). 
This process involves generating varied textual descriptions of the same meaning, while preserving both main contents and detail attributes unchanged.
The details of utilized LLMs and prompts used in the two steps are shown in Appendix~\ref{vld_data_gen_detail}.

\noindent\textbf{(b) Video Instruction Tuning Data.}
Alongside with the emergence of LMMs, extensive video instruction tuning datasets have also been publicly available.
For example, VideoInstruct100K~\cite{Maaz2023VideoChatGPT} contains question-answer pairs related to video summarization,
description-based question-answers,
and creative/generative question-answers.
VideoChat~\cite{li2023videochat} provides a rich dataset featuring elaborate video descriptions and dialogues, enhancing data variety by embracing temporal and causal aspects within video instructions.
These datasets were originally crafted to train a genre-independent video understanding model, rather than to curate video descriptions. Consequently, our method includes employing LLMs for \textit{Data Filtering} to exclude samples extraneous to video descriptions. 
We employ prompt engineering and also provide some demonstration examples to aid LLMs in achieving better affects.
Finally, the \textit{Desc. Rewrite} step is also conducted.
The details of utilized LLMs and prompts are shown in Appendix~\ref{vld_data_gen_detail}.

\noindent\textbf{(c) Available Video \& Long-Description Data.}
As previously mentioned, existing datasets pairing videos with long text descriptions are often limited by either the quantity or the domains/genres of videos.
In this regard, we perform the \textit{Data Sampling} operation over these datasets.
Specifically, 57.8K video clips of gaming and city/scenic exploration scenarios in MiraData~\cite{miradata} are all included in VILD.
50K long captions describing natural landscape are randomly sampled from Open-Sora-Dataset~\cite{opensoradata}.
The \textit{Desc. Rewrite} step is also involved at the end.

\noindent\textbf{(d) Raw Video Data.}
In order to further expand the amount of training data, we leverage LMMs and LLMs to generate long descriptions given raw videos (some combined with corresponding short captions). 
An optional \textit{Short Video Desc. Generation} step is required using off-the-shelf  models~\cite{BLIP2,tag2text,zhang2023videollama,videoblip} if there are no short captions available. 
For computation efficiency, we randomly sample over 2M video clips with high-quality short captions generated by a number of teacher models and a fine-tuned caption selection model from Panda-70M~\cite{chen2024panda}.
Then, we sample \textit{k} (\textit{k}=3 in our setting) frames from each video clip at equal intervals as key-frames and use LMMs to annotate them with long descriptions.
We do not conduct this for each frame, as it would be extremely time-consuming and laborious.
Next, given short description of the whole video and long descriptions of its key-frames, we ask LLMs to integrate them into long description of the whole video. 
The assistance of short video description can alleviate the hallucinations present in frame descriptions. 
Our findings have also reached a consensus with existing studies~\cite{viclip,wang2024internvideo2} that directly using video LMMs~\cite{li2023videochat,Maaz2023VideoChatGPT} to describe videos for long captions can lead to sub-optimal results.
The details of utilized LLMs/LMMs and prompts are shown in Appendix~\ref{vld_data_gen_detail}.

Finally, the \textit{Post-Processing} step is performed. 
NSFW examples are filtered out. Next, we utilize ViCLIP~\cite{viclip} and Long-CLIP~\cite{longclip}  to filter out low-quality examples with average video-text similarity smaller than 0.20. 
We finally collect over 2M video \& long-description data pairs as our VILD dataset for model pre-training. More detailed comparisons of data statistics information are shown in Appendix~\ref{appendix_vld_static}.

\subsection{Text-similarity-guided Primary Component Matching (TCPM)}\label{dpcm}

The vanilla pre-training of CLIP models consumes vision-text pairs ($v$, $t$) as inputs. $v$ can be images or videos.
It makes no assumptions on specific single-modal encoder architectures. 
Given a vision encoder $E_v$ and a text encoder $E_t$,
single-modal features are first extracted as $f_v=E_v(v)$, $f_t=E_t(t)$.
Then, contrastive learning with the InfoNCE~\cite{infonce} loss typically is employed to learn the correspondence between vision and text.
In particular, this can be formulated as:

\begin{small}
\begin{align}
\mathcal{L}_{\mathrm{CL}}(f_t, f_v) &= \frac{1}{2N} \sum\nolimits_N \mathcal{L}^{f_t\rightarrow f_v}_{\text{InfoNCE}}+\mathcal{L}^{f_v\rightarrow f_t}_{\text{InfoNCE}},
\end{align}\end{small}
where $N$ is the batch size and 
\begin{small}
\begin{align}
\mathcal{L}^{f_t\rightarrow f_v}_{\text{InfoNCE}} &= -\log \frac{\exp(sim(f_t, f_v^+ ) / \tau)}{\sum_{f_v\in \{f_v^+,f_v^-\}} \exp(sim(f_t, f_v) / \tau)},
\end{align}
\end{small}and vise versa. Here, $\tau$ is the temperature hyper-parameter,
$sim$ is the cosine similarity calculation,
$f_v^+$ is the \textit{positive} vision feature which is paired with the text feature $f_t$, 
and $f_v^-$ are \textit{negative} vision features that are formed by other unpaired images/videos in the current training batch. 

To extend the long-description understanding capacity of CLIP models, Long-CLIP~\cite{longclip} is proposed to use primary component matching for image CLIPs.
Given the short description, the long description, and the vision input (\textit{st}, \textit{lt}, \textit{v}),
the loss function is formulated as:
\begin{align}
\mathcal{L}= \mathcal{L}_{\mathrm{CL}}(f_{lt}, f_v)+\alpha_1 \mathcal{L}_{\mathrm{CL}}(f_{st}, f'_v),
\end{align}
where $\alpha_1$ is the ratio hyper-parameter and $f'_v= \mathrm{PCE}(f_v, \ 32)$.
Here, PCE is short for primary component extraction that consists of the component-decomposition function $\mathcal{F}$ (which decomposes the feature into vectors of different attributes and their importance), 
the component-filtration function $\mathcal{E}$ (which filters out less important attributes), and the component reconstruction function $\mathcal{F}^{-1}$ (which reconstructs the feature).
In the implementation of $\mathcal{E}$, Long-CLIP selects the most important 32 attributes as the retained ones.

\begin{figure}[t]
\centering
\includegraphics[width=0.49\textwidth]{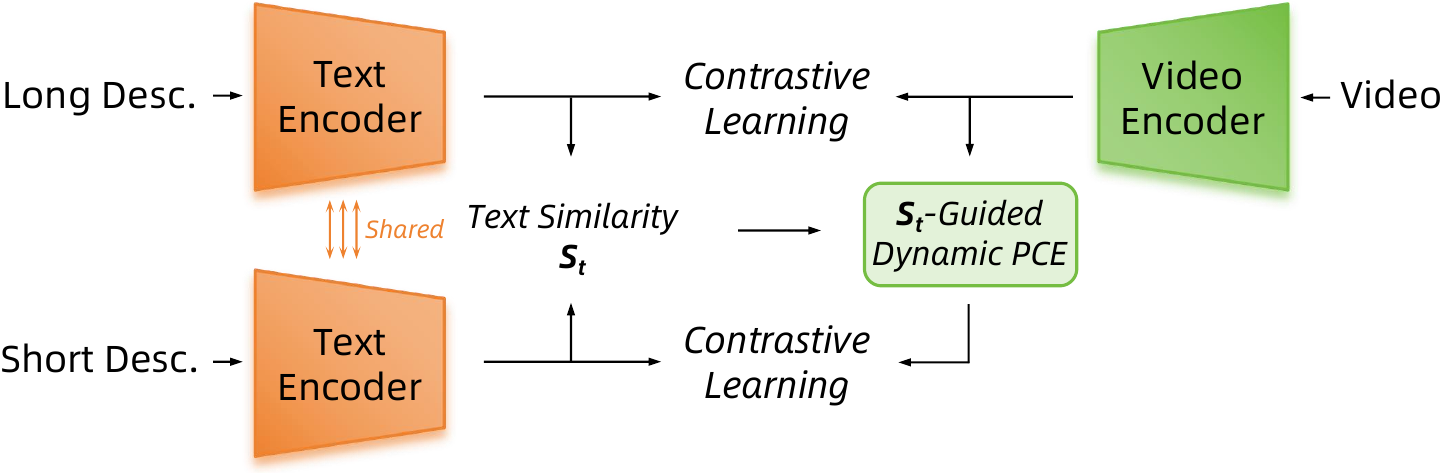}
\caption{The proposed text-similarity-guided primary component matching (TPCM) technique. 
} 
\label{image-tpcm}
\end{figure}

However, when extending this technique for video pre-training, we have found that since videos usually contain richer contents and more details than images, this fixed strategy cannot dynamically adapt to the severe distribution changes of high-dimensional feature spaces of video CLIPs during learning (shown in Fig.~\ref{img:pca_dim}).
In this regard, we propose to use the cosine text similarity between $lt$ and $st$ as a signal to guide the PCE process, as shown in Fig.~\ref{image-tpcm}. Therefore, we re-write $\Hat{f}_v$ as follows:
\begin{align}
\Hat{f}_v = \mathrm{PCE}(f_v,\  \mathcal{G}(sim(f_{lt}, \ f_{st}))),
\end{align}
where $\mathcal{G}$ represents that we preserve the attributes in descending order of importance until the similarity between $\Hat{f}_v$ and $f_v$ reaches the similarity between $lt$ and $st$.

\subsection{Two Description Ranking Tasks}\label{drt}
\begin{figure}[t!]
\centering

\includegraphics[width=7.5cm]{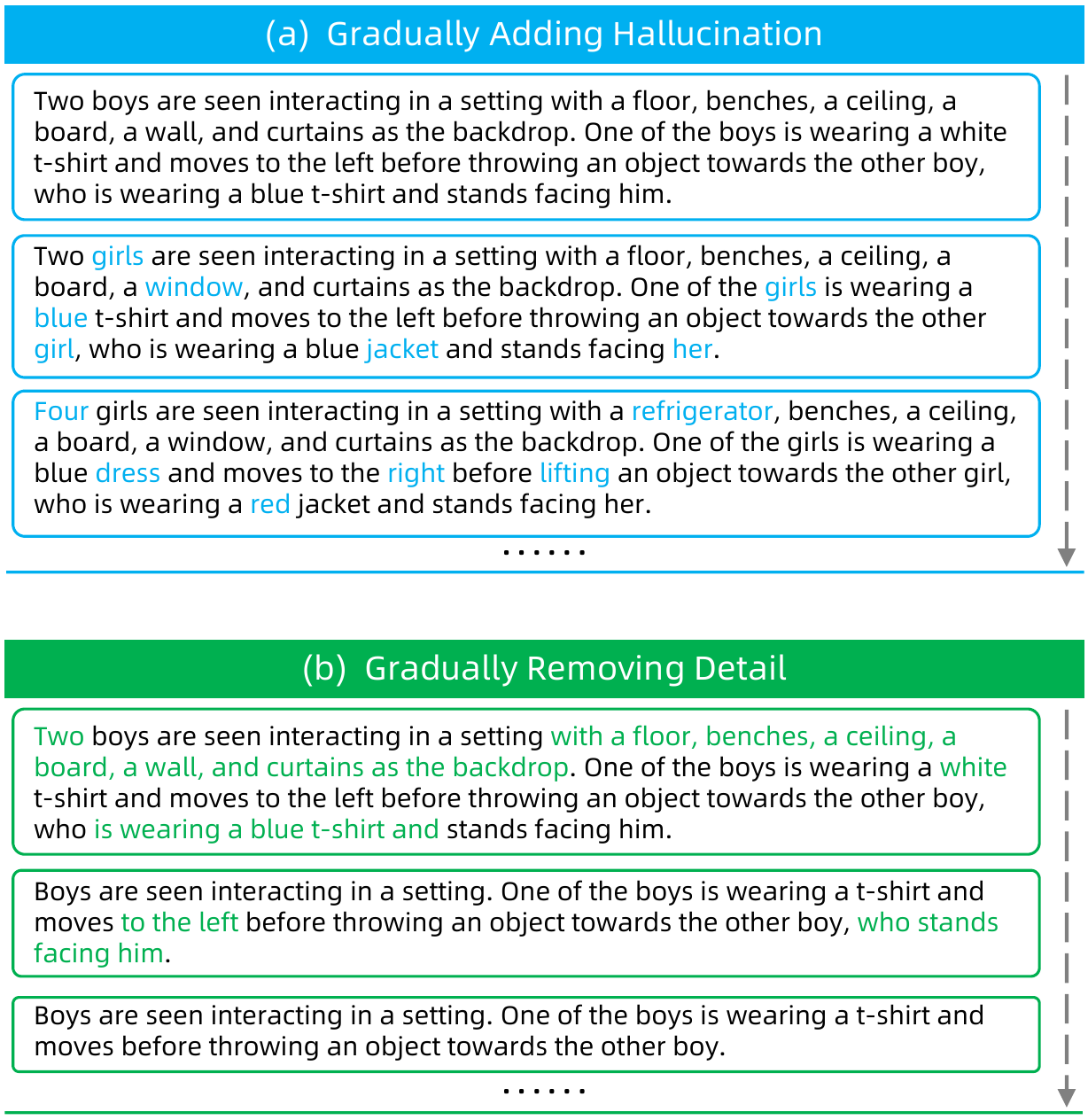}
\caption{Examples of text samples generated for (a) hallucination-aware and (b) detail-aware description ranking tasks.
\textit{Blue} and \textit{green} words refer to replaced hallucination content and detailed content to be deleted, respectively. Best viewed in color.
}
\label{image-drt}
\end{figure}

We posit that video CLIP models designed to comprehend long descriptions should inherently exhibit two characteristics: given a video and its associated descriptions, the model should assign a higher score to descriptions (1) with richer and more precise contexts and (2) that are more accurate and less prone to hallucinations, given an equivalent level of detail. 
To realize these principles, we introduce two novel tasks: Detail-aware Description Ranking (DDR) and Hallucination-aware Description Ranking (HDR) to address the respective attributes. Our preparatory steps involve employing syntactic analysis tools, such as NLTK~\cite{nltk} and spaCy~\cite{spacy}, to execute part-of-speech tagging and parse syntactic structures within the long-description ground truths.

\begin{figure*}[t]
\centering
\includegraphics[width=0.99\textwidth]{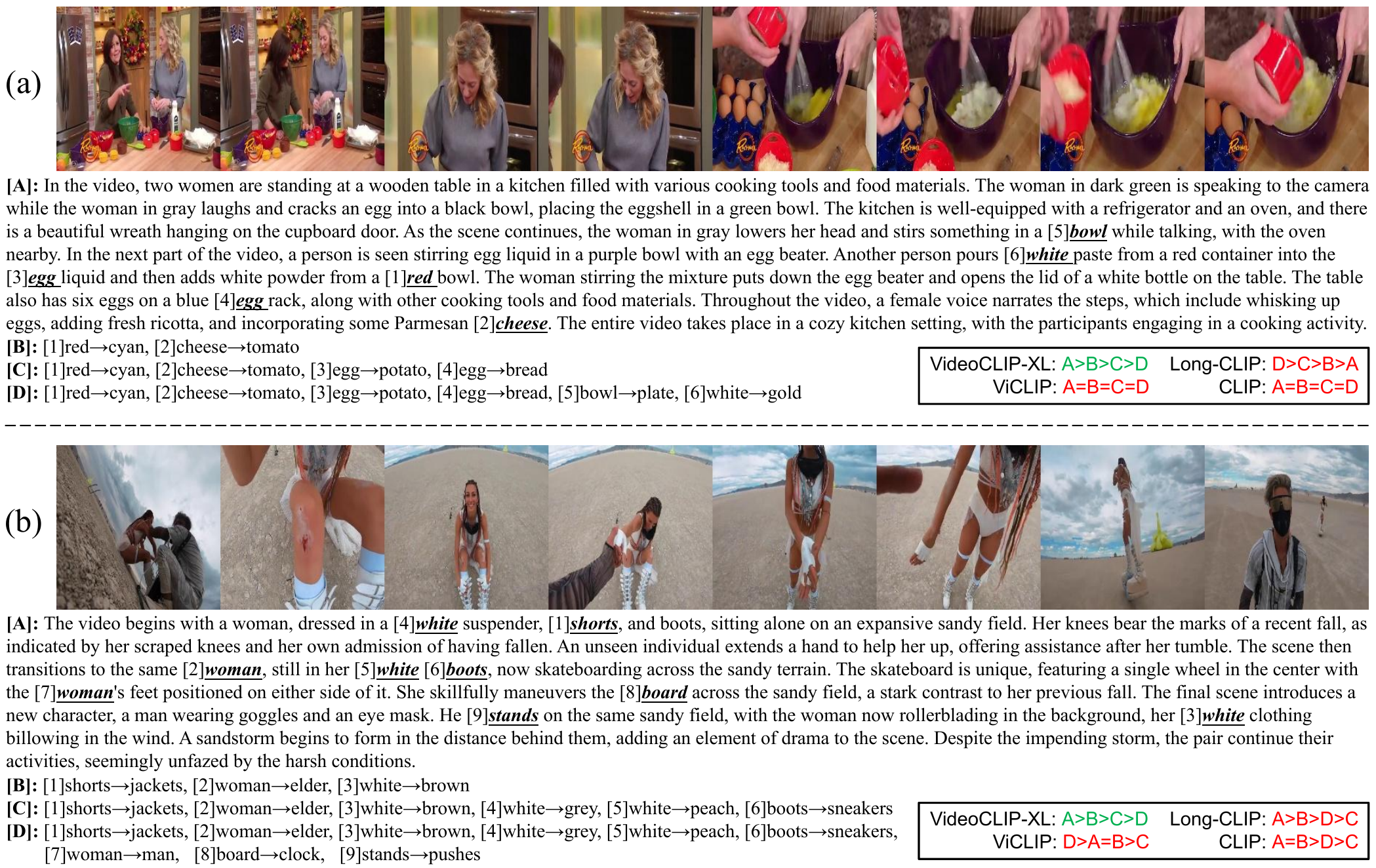} 
\caption{Qualitative examples on our LVDR benchmark.
We calculate the cosine similarities between different long descriptions and the same video using video CLIP models. 
Descriptions are sorted based on these similarities in descending order after retaining 8 decimal places.
}
\label{img:lvdr_example}
\end{figure*}

Subsequently, we synthesize multiple description candidates for each video to facilitate DDR and HDR training. As illustrated in Fig.~\ref{image-drt}(a), in each step, we selectively replace specific words (nouns, numerals, colors, or terms related to direction, verbs) with their semantically disparate counterparts within the same syntactic category (e.g., \textit{boys} to \textit{girls}, \textit{white} to \textit{blue}, \textit{throwing} to \textit{lifting}), and perform this replacement $m-1$ times. This method yields a series of progressively hallucinated descriptions, denoted as $\mathbf{t^H}=\{t^H_1, t^H_2, \ldots, t^H_{m}\}$.
Analogously, as depicted in Fig.~\ref{image-drt}(b), each step involves randomly excising sub-sentences, adjectives, numerals, or dependency parse sub-trees from the current description. This process recursively generates $m-1$ sequentially less detailed descriptions for each video, expressed as $\mathbf{t^D}=\{t^D_1, t^D_2, \ldots, t^D_{m}\}$.

For $\mathbf{t^H}$ or $\mathbf{t^D}$, given the same corresponding video, 
we hope that the model can generate a higher similarity score for the description appearing earlier in the sequence. For example, for the DDR task, we formulate the loss function as follows:

\begin{small}
\begin{align}
 \mathcal{L}_{\mathrm{DDR}}= \frac{1}{\frac{m(m-1)}{2}} \sum_{i=1}^{m-1} \sum_{j=i+1}^m \mathrm{ReLU}(-(\Delta^D_{i,j}-\alpha_{D})),
\end{align}
\end{small}where $\alpha_{D}$ is the similarity difference gap and
\begin{align}
\Delta^D_{i,j} &= sim(f_{t^D_i},f_v) - sim(f_{t^D_j},f_v).
\end{align}
The intuition behind this learning objective comes from the requirement for the model to be able to differentiate between various descriptions with the minimum distinction degree $\alpha_D$.
Similarly, for HDR, we have the loss function:

\begin{small}
\begin{align}
\mathcal{L}_{\mathrm{HDR}} &= \frac{1}{\frac{m(m-1)}{2}} \sum_{i=1}^{m-1} \sum_{j=i+1}^m \mathrm{ReLU}(-(\Delta^H_{i,j}-\alpha_{H})).
\end{align}
\end{small}
The total loss of our pre-training process is:
\begin{small}
\begin{align}
\mathcal{L}= &\mathcal{L}_{\mathrm{CL}}(f_{lt}, f_v)+\alpha_1 \mathcal{L}_{\mathrm{CL}}(f_{st}, f'_v)\nonumber+\\
&\alpha_2 \mathcal{L}_{\mathrm{DDR}}+\alpha_3 \mathcal{L}_{\mathrm{HDR}},\label{equal:total_loss}
\end{align}
\end{small}
where $\alpha_2$ and $\alpha_3$ are balancing hyper-parameters.

\begin{table*}[t]
\centering
\begin{small}
\begin{tabular}{@{}lcccccccccccc@{}}
\toprule
\multirow{2}{*}{Method} & \multicolumn{2}{c}{MSR-VTT} & \multicolumn{2}{c}{LSMDC} & \multicolumn{2}{c}{DiDeMo} & \multicolumn{2}{c}{MSVD} & \multicolumn{2}{c}{ActivityNet}  & \multicolumn{2}{c}{Avg.}\\
\cmidrule(l){2-13}
& T2V  & V2T  & T2V  & V2T  & T2V  & V2T  & T2V  & V2T  & T2V  & V2T   & T2V  & V2T  \\ 
\midrule
CLIP~\cite{clip}      & 30.4 & 24.2 & 13.9 & 11.9 & 12.7 & 18.7 & 40.5 & 57.2 & 9.1 & 13.2 & 21.3 & 25.0 \\
VideoCLIP~\cite{videoclip} & 10.4  & - & -& -& 16.6 & -& -& -& -& -& -& - \\
CLIP4Clip~\cite{clip4clip} & 32.0 & - & 15.1 & - & - & - & 38.5 & - & - & -& -& -  \\
ViCLIP~\cite{viclip}      & 42.4 & 41.3 & 20.1 & 16.9 & 38.7 & 39.1 & 49.1 & 75.1 & 32.1 & 31.4 & 36.5 & 40.8  \\
\midrule
\textbf{VideoCLIP-XL (Ours)} & \textbf{50.1}  & \textbf{49.9}     & \textbf{22.8}  & \textbf{24.6}  & \textbf{47.7}  & \textbf{47.9}  & \textbf{51.9}  & \textbf{76.7}  & \textbf{46.4$^\ddag$}  & \textbf{48.1$^\ddag$} &\textbf{43.8} &\textbf{49.5}  \\
\bottomrule
\end{tabular}%
\end{small}
\caption{R@1 scores of zero-shot text-video retrieval on MSR-VTT~\cite{xu2016msr}, LSMDC~\cite{lsmdc}, DiDeMo~\cite{didemo}, MSVD~\cite{msvd}, and ActivityNet~\cite{caba2015activitynet}. T2V and V2T are short for text-to-video and video-to-text, hereinafter the same.
$^\ddag$Due to the high overlap between the videos in ActivityNet and VideoInstruct100K~\cite{Maaz2023VideoChatGPT}, the latter is excluded from the pre-training data of our model tested on the former.  
}
\label{tab:zs_traditional_data}
\end{table*}

\begin{table*}[t]
\centering
\begin{small}
\begin{tabular}{@{}lcccccccccccc@{}}
\toprule
\multirow{2}{*}{Method} & \multicolumn{2}{c}{MSR-VTT} & \multicolumn{2}{c}{LSMDC} & \multicolumn{2}{c}{DiDeMo} & \multicolumn{2}{c}{MSVD} & \multicolumn{2}{c}{ActivityNet}  & \multicolumn{2}{c}{Avg.}\\
\cmidrule(l){2-13}
& T2V  & V2T  & T2V  & V2T  & T2V  & V2T  & T2V  & V2T  & T2V  & V2T  & T2V  & V2T   \\ 
\midrule
CLIP~\cite{clip}      & 38.2 & 38.7 & 22.5 & 22.6 & 32.2 & 33.9 & 52.3 & 69.9 & 26.1 & 26.9 & 34.3 & 38.4 \\
VideoCLIP~\cite{videoclip} & 30.9  & -& -& -& -& -& -& -& -& - & -& - \\
CLIP4Clip~\cite{clip4clip} & 45.6 & 45.9 & 24.3 & 23.8 & 43.0 & 43.6 & 45.2 & 48.4 & 40.3 & 41.6 & 39.7& 40.7 \\
ViCLIP~\cite{viclip}      & 52.5 & 51.8 & 33.0 & 32.5 & 49.4 & 50.2 & 53.1 & 79.0 & 49.8 & 48.1 & 47.6 & 52.3 \\
\midrule
\textbf{VideoCLIP-XL (Ours)} & \textbf{57.0} & \textbf{56.6}  & \textbf{34.2}  & \textbf{32.6}  & \textbf{62.3}  & \textbf{62.7}  & \textbf{55.6}  & \textbf{81.4}  & \textbf{58.4$^\ddag$}  & \textbf{59.2$^\ddag$} & \textbf{53.5}  & \textbf{58.5}   \\
\bottomrule
\end{tabular}%
\end{small}
\caption{R@1 scores of fine-tuned text-video retrieval on MSR-VTT, LSMDC, DiDeMo, MSVD, and ActivityNet.  $^\ddag$Due to the high overlap between the videos in ActivityNet and VideoInstruct100K~\cite{Maaz2023VideoChatGPT}, the latter is excluded from the pre-training data of our model tested on the former.
}
\label{tab:ft_traditional_data}
\end{table*}

\subsection{The New LVDR Benchmark}\label{lvdr}
Hallucination is ubiquitous in contemporary LLMs and LMMs~\cite{hallucination}.
Given a video, the video CLIP model with the ability to understand long texts should naturally possess the discernment to distinguish between correct and erroneous texts in long descriptions.
To better evaluate such ability, we propose the Long Video Description Ranking (LVDR) benchmark.
We first randomly sample 2K video \& long-description pairs from Shot2Story~\cite{shot2story}.
Then, we perform a synthesis process similar to Fig.~\ref{image-drt}(a), iterating $p-1$ times and altering $q$ words during each iteration, and resulting in totally $p$ descriptions with gradually increasing degrees of hallucination.
We denote such a subset as $p\times q$ and construct five subsets as \{$4\times 1$, $4\times 2$, $4\times 3$, $4\times 4$, $4\times 5$\}.
Each distinct subset 
is manually reviewed to avoid inappropriate replacement. 
Representative examples are provided in Fig.~\ref{img:lvdr_example}.
Based on our analysis, a better model needs to be able to correctly rank these descriptions in descending order of similarity given the video.
Thus, we also design the evaluation criterion named ranking score (RS) which can be formulated as:

\begin{small}
\begin{align}
\mathrm{RS} = \frac{100}{\frac{m(m-1)}{2}} \sum_{i=1}^{m-1} \sum_{j=i+1}^m \mathds{1}(sim(f_{t_i},f_v) > sim(f_{t_j},f_v)).
\end{align}
\end{small}Here, $\mathds{1}$ is the indicator function.

\section{Experiments}

\subsection{Implementation Detail}

We adopt the model structure of CLIP~\cite{clip} with ViT-L/14 and leverage spatio-temporal attention in the video encoder with the weight initialization from ViCLIP~\cite{viclip}.
We further pre-train VideoCLIP-XL on our VILD dataset for 2 epochs.
All experiments are implemented in PyTorch and run on  NVIDIA Tesla A100-80G GPUs. 
More experimental details are given in Appendix~\ref{experimental_setting_detail}.

\subsection{Performance Comparison}

\begin{table*}[t]
\centering
\resizebox{0.99\textwidth}{!}{%
\begin{tabular}{@{}lccccc@{}}
\toprule
\multirow{2}{*}{Method} & \multicolumn{5}{c}{LVDR Benchmark}  \\
\cmidrule(l){2-6}
& $4\times 1$  & $4\times 2$  & $4\times 3$  & $4\times 4$ & $4\times 5$ \\ 
\midrule
CLIP~\cite{clip}      & 30.12/27.83/18.76 & 47.97/38.46/31.61 & 57.00/43.61/39.31 &65.31/50.53/48.78 & 69.58/53.17/53.31  \\
ViCLIP~\cite{viclip}    &18.95/24.93/18.40 &33.63/41.57/32.43&44.47/52.06/43.10& 52.43/58.81/50.46&58.61/62.10/55.55    \\
Long-CLIP~\cite{longclip}   & 70.07/49.67/51.11 & 81.66/65.45/68.78 &  86.49/73.91/77.54 & 89.84/79.90/83.32& 91.70/83.60/86.50 \\
\midrule
\textbf{VideoCLIP-XL (Ours)} & \textbf{80.32}/\textbf{70.93}/\textbf{72.06} & \textbf{90.72}/\textbf{83.70}/\textbf{86.11} & \textbf{94.41}/\textbf{89.87}/\textbf{91.92} & \textbf{95.87}/\textbf{91.93}/\textbf{93.49}  & \textbf{96.99}/\textbf{94.17}/\textbf{95.47}  \\
\bottomrule
\end{tabular}%
}
\caption{Ranking score (RS)/Kendall’s tau (KT)/Spearman rank-order correlation coefficient (SC) of long video description ranking on the proposed LVDR benchmark.}
\label{tab:lvdr}
\end{table*}
We compare VideoCLIP-XL with strong competitors in three different downstream tasks: text-video retrieval on traditional benchmarks, text-video retrieval on long-description benchmarks, and description ranking on our LVDR benchmark.

Results on traditional benchmarks for text-video retrieval are shown in Tab.~\ref{tab:zs_traditional_data} and~\ref{tab:ft_traditional_data}.
We can find that, VideoCLIP-XL exhibits superior performance on all benchmarks compared with other video CLIP models under both zero-shot and fine-tuning settings.
For example, VideoCLIP-XL outperforms the previous state-of-the-art ViCLIP, with an improvement of  +7.7/+8.6 T2V/V2T zero-shot R@1 scores and +4.5/+4.8 T2V/V2T fine-tuning R@1 scores on MSR-VTT. 
It is worth noting that, although our method mainly focuses on learning fine-grained features in videos and texts, 
its effective training strategy can also result in significant improvements on all benchmarks, regardless of whether the texts are detailed or not.

As in Tab.~\ref{tab:zs_shot2story_data}, VideoCLIP-XL also surpasses other competitors significantly on Shot2Story under the long description setting. 
In Shot2Story, each video clip consists of multiple video shots which switch between different scenes to express the same main event.
This requires the model to have the ability to fully understand mainline activity from multiple complex scenarios. Performances demonstrate that our method exhibits significant advantages whether using the whole video clip (Shot2Story-W) or each shot (Shot2Story-S) as an individual for the text-video retrieval task.

\begin{table}[t]
\centering
\begin{small}
\setlength{\tabcolsep}{0.5mm}{
\begin{tabular}{@{}lcccc@{}}
\toprule
\multirow{2}{*}{Method} & \multicolumn{2}{c}{Shot2Story-W} & \multicolumn{2}{c}{Shot2Story-S}  \\
\cmidrule(l){2-5}
& T2V  & V2T  & T2V  & V2T \\ 
\midrule
CLIP~\cite{clip}      &65.80 & 66.00 &45.40  & 45.35  \\
ViCLIP~\cite{viclip}    & 37.53 & 37.71 &48.44 & 46.17   \\
Long-CLIP~\cite{longclip}   &74.74  &80.59 &47.70 & 43.14  \\
\midrule
\textbf{VideoCLIP-XL (Ours)} & \textbf{95.28} & \textbf{94.73} & \textbf{70.30} &  \textbf{67.79}  \\
\bottomrule
\end{tabular}
}
\end{small}
\caption{R@1 of text-video retrieval on Shot2Story~\cite{shot2story} with long video descriptions.}
\label{tab:zs_shot2story_data}
\end{table}

\begin{table}[t]
\centering
\setlength{\tabcolsep}{2.2mm}{
\begin{small}
\begin{tabular}{@{}clccc@{}}
\toprule
\# & Pre-Training Data & \tabincell{c}{MLDMA\\(R@1)}  & \tabincell{c}{S2S\\(R@1)}   & \tabincell{c}{LVDR\\(RS)}  \\
\midrule
1 & Part (a)(b)(c) of VILD   &45.52 & 80.37   &  90.97   \\
2 & Part (d) of VILD   &44.54  & 78.77   &   89.23 \\
3 & Full VILD & \textbf{46.61} & \textbf{82.03} &  \textbf{91.67} \\
\bottomrule
\end{tabular}
\end{small}
}
\\(a)\\
\bigskip
\setlength{\tabcolsep}{2.0mm}{
\begin{small}
\begin{tabular}{@{}ccccccc@{}}
\toprule
\# & TPCM & DDR & HDR & \tabincell{c}{MLDMA\\(R@1)}  & \tabincell{c}{S2S\\(R@1)}   & \tabincell{c}{LVDR\\(RS)} \\
\midrule
1 & \multicolumn{3}{l}{\tabincell{l}{Baseline\\~\cite{longclip}}}  & 45.62 &81.47 &  84.87 \\ \midrule
2 & $\checkmark$ & &  &46.06 &\textbf{82.03} & 84.87  \\
3 & $\checkmark$ & $\checkmark$ &  &46.58 & \textbf{82.03} & 86.07  \\
4 & $\checkmark$ & & $\checkmark$  & 46.07 & \textbf{82.03} & 91.42 \\
5 & $\checkmark$ & $\checkmark$ & $\checkmark$  & \textbf{46.61} & \textbf{82.03} &   \textbf{91.67} \\
\bottomrule
\end{tabular}
\end{small}
}
\\(b)\\
\caption{Ablation study for components of our method.
MLDMA indicates the averaged zero-shot text-video retrieval R@1 score of benchmarks in Tab.~\ref{tab:zs_traditional_data}.
S2S is short for Shot2Story.
}
\label{tab:ablation}
\end{table}

The results on our LVDR benchmark are shown in Tab.~\ref{tab:lvdr}.
VideoCLIP-XL has a stronger identification ability compared with competitors to perceive inaccurate content in long video descriptions and assign them lower similarity scores. For example, under the $4 \times 1$ setting where \textit{only 1 original word is randomly replaced with a wrong one} between adjacent generated descriptions, our model can surpass Long-CLIP (which focuses on long text understanding for images) with +10.25 ranking score.
We can also observe that as the level of single-step hallucination increases from shallow to deep ($4\times 1$ to $4\times 5$), the video CLIP models can naturally distinguish different long video descriptions better.

\subsection{Ablation Study}
In this subsection, we aim to explore the effectiveness of each component in our method.

As shown in Fig.~\ref{image-vld}, our VILD pre-training dataset is formed by the aggregation of four parts from different data sources.
For parts (a)(b)(c), the data resource often utilizes the powerful GPT-4V~\cite{GPT4v} or human efforts to generate the text information
before our LLM-based steps.
While for part (d), we use open-source LLMs for generating long descriptions from raw videos.
The results in Tab.~\ref{tab:ablation}(a) show the data effectiveness.
Although the effect of using open-source LLMs for automated data synthesis can naturally lag behind GPT-4V/human efforts by a margin, 
it can still achieve state-of-the-art performance compared with existing competitors. 
In addition, adding (d) on top of (a)(b)(c) can further result in obvious improvements. 
This also demonstrates the effectiveness of our data synthesis pipeline.

\begin{figure}[t]
\centering
\includegraphics[width=0.4\textwidth]{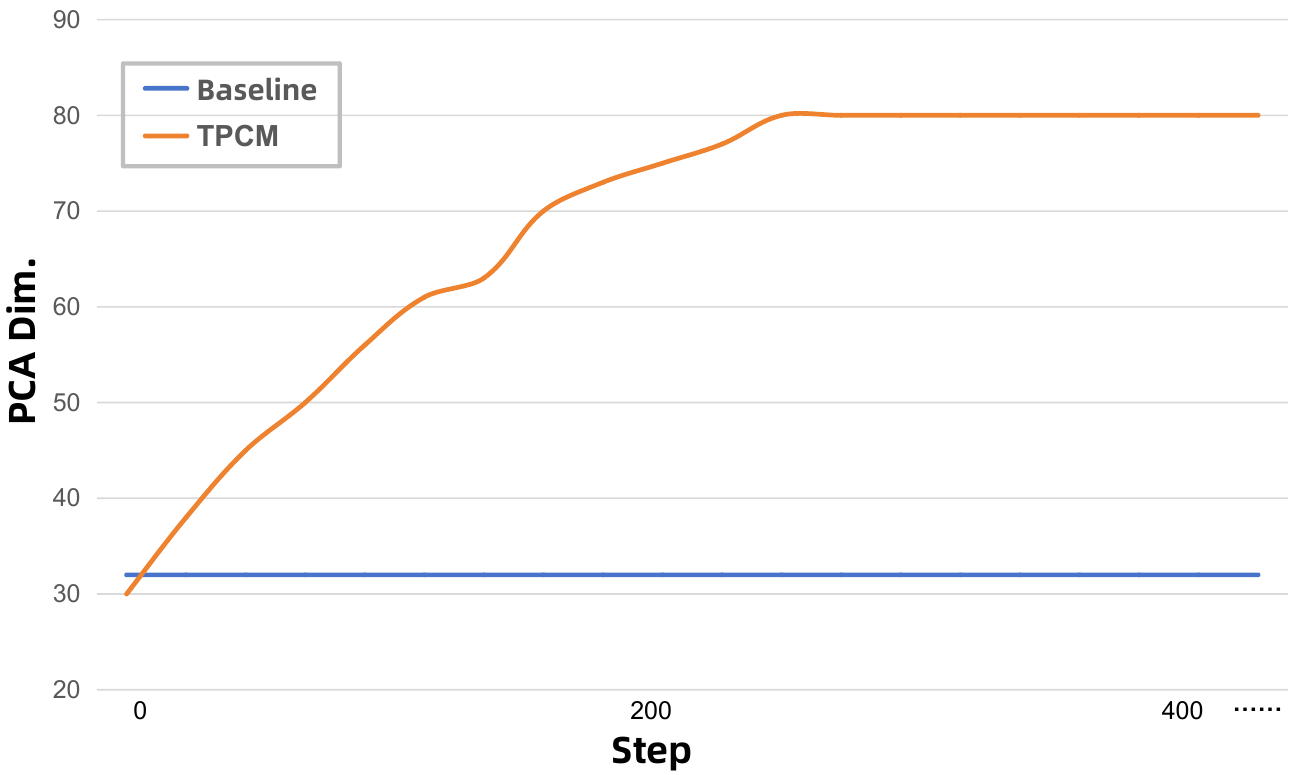} 
\caption{
TPCM can dynamically adjust the dimension of the attribute vectors that need to be retained during pre-training.
Dim. is short for dimension.
}
\label{img:pca_dim}
\end{figure}

As shown in Tab.~\ref{tab:ablation}(b) \#2 v.s. \#1,
TPCM can achieve +0.44 R@1 gain on traditional text-video retrieval datasets and +0.56 R@1 gain on Shot2Story.
In addition, it can dynamically modify the feature space distribution during pre-training, which is reflected in the increase of PCA dimension,
as shown in Fig.~\ref{img:pca_dim}.

The effectiveness of DDR and HDR can also be found in Tab.~\ref{tab:ablation}(b).
Compared \#3 with \#2,
DDR achieves a +0.52 R@1 gain on traditional benchmarks and +1.20 RS gain on LVDR.
As for HDR, compared \#4 with \#2, it obtains +6.55 RS gain on LVDR.
Furthermore, conducting both tasks together is more effective than using
either one alone on MLDMA and LVDR, as shown in \#5 v.s. \#2.

\section{Conclusion}
In this paper, we propose \textbf{VideoCLIP-XL}, a video CLIP model with long-description capability.
We establish an automatic data collection system to gather our VILD dataset, 
and propose TPCM to better learn the distribution of feature space during pre-training while expanding the long-description capability.
We also introduce two new tasks namely DDR and HDR for further understanding improvement.
Our LVDR benchmark is helpful for evaluating the long-description capability more comprehensively.
Extensive experimental results have demonstrated the effectiveness of our method.

For future research, we plan to refine the pre-training methodology and increase the amount of data and model size for further improvement.
We will also attempt to integrate the architecture of cross-encoders and LLMs into our method.
% \newpage

\section*{Limitations}
Although VideoCLIP-XL is trained to enable long-description understanding capacity, 
limited by the amount of pre-training data
and the feature extraction capability of single-modal encoders, there is still room for improvement.
The scale, quality, and variety of data can be further extended,
and the model structure and model size of feature extractors can also be scaled up.
The application of our method in the structures of cross-encoders and LLMs is also worth exploring.
These improvements are left to our subsequent work.

\section*{Ethical Considerations}
The techniques for training the VideoCLIP-XL model
presented in this work are fully methodological,
thereby there are no direct negative social impacts
of our method. Additionally, we have filtered out
NSFW examples from our pre-training data to ensure
that the seen contents are suitable for public distribution.

\section*{Acknowledgements}
This research is supported in part by National Natural Science Foundation of China (Grant No.: 62441604, 62476093). 
It is also partially supported by Alibaba Cloud Computing, through Research Talent Program with South China University of Technology.

\bibliography{custom}

\appendix

\section{Appendix}

\subsection{Details of VILD Data Generation}
\label{vld_data_gen_detail}
During the data generation for VILD, 
we leverage Qwen1.5-72B-Chat~\cite{qwen} in LLM-based steps and LLaVA-v1.6-34B~\cite{liu2024llavanext} in LMM-based steps. All prompts we used are listed as follows:

\textbf{[\textit{Desc. Aggregation}]}

\textit{``The following are descriptions of the subjects or background in a video. Please organize them together into a single description of the entire video. Do not omit any content, nor add any new content that is not included or uncertain.}

\textit{\{examples\}}

\textit{Descriptions: \{individual-level descriptions\}}

\textit{Output:}''

\textbf{[\textit{Desc. Rewrite}]}

\textit{``The following is a video description. Please output a rewritten version. Do not omit any content, nor add any new content that is not included or uncertain.}

\textit{\{examples\}}

\textit{Description: \{input description\}}

\textit{Output:}''

\textbf{[\textit{Data Filtering}]}

\textit{``Determine if the following conversation is talking about the overall/comprehensive-level description/content of a video. If yes, output Yes; otherwise, output No.}

\textit{\{examples\}}

\textit{Conversation: \{input conversation\}}

\textit{Output:}''

\textbf{[\textit{Long Frame Desc. Generation}]}

\textit{``Precisely describe this image.}''

\textbf{[\textit{Long Video Desc. Generation}]}

\textit{``We will provide a description of a video and some frame descriptions of it. Directly output an enriched video description according to them. Remove repetitive contents. Do not describe any content that is uncertain or not included. Do not describe individual frames. Do not describe specific subjects, use generic words instead.}

\textit{\{examples\}}

\textit{Video Description: \{short video description\}}

\textit{Frame Descriptions: \{long frame descriptions\}}

\textit{Output:}''

\subsection{Details of Data Statistics}
\label{appendix_vld_static}
More detailed comparisons of data statistics information are shown in Tab.~\ref{tab:appendix_data_static}.
\begin{table*}[t]
\centering
\resizebox{0.97\textwidth}{!}{%
\begin{tabular}{@{}lcccccccccccc@{}}
\toprule
\textbf{Dataset}  &	\textbf{Year}	 &\textbf{Caption Source}	 &\textbf{Domain}	 &\textbf{Video Num.}  &	\textbf{Avg. Video Len.}	 & \textbf{Avg. Text Len.} \\
\midrule
HowTo100M~\cite{miech2019howto100m}	&2019	&ASR&	Open&	136M&	3.6s	&4.0 words\\
HD-VILA-100M~\cite{hdvila100m} &	2022	&ASR&	Open	&103M	&13.4s	&32.5 words\\
MSVD~\cite{msvd} &	2011	&Manual&	Open&	1970&	9.7s	&8.7 words\\
LSMDC~\cite{lsmdc} &	2015&	Manual&	Movie	&118K	&4.8s	&7.0 words\\
MSR-VTT~\cite{xu2016msr}	&2016&	Manual&	Open	&10K&	15.0s	&9.3 words\\
DiDeMo~\cite{didemo}	&2017&	Manual&	Flickr	&27K&	6.9s	&8.0 words\\
ActivityNet~\cite{caba2015activitynet}	&2017&	Manual	&Action&	100K	&36.0s	&13.5 words\\
YouCook2~\cite{youcook} &	2018&	Manual&	Cooking	&14K&	19.6s	&8.8 words\\
VATEX~\cite{wang2019vatex} &2019&	Manual&	Open	&41K&	~10.0s	&15.2 words\\
Panda-70M~\cite{chen2024panda} &	2024	&Automatic	&Open	&70.8M	&8.5s	&13.2 words\\ \midrule
\textbf{VILD (Ours)}	&2024	&Automatic	&Open&	2.1M	& 15.4s 	&74.2 words\\
\textbf{LVDR (Ours)}	&2024	&Automatic+Manual&	Open	&2K& 17.5s	 &	230.7 words\\
\bottomrule
\end{tabular}%
}\caption{Comparison of data statistics information.}
\label{tab:appendix_data_static}
\end{table*}

\subsection{Details of Experimental Settings}
\label{experimental_setting_detail}
We sample 8 frames for each video during pre-training.
Stretching of the vanilla absolute positional embedding from 77 to 248 is also applied following~\cite{longclip}.
During pre-training, we set the batch size 1664, warm-up steps 200, weight decay 0.02, and max learning rate 4e-6.
The learning rate decreases in a cosine schedule after warm-up.
$\alpha_1$, $\alpha_2$, $\alpha_3$, $\alpha_{D}$, and $\alpha_{H}$ are empirically set as 0.1, 1.0, 10.0, 0.0, and 0.0 respectively.
$m$ in the DDR and HDR tasks is set as 5.

During pre-training, as shown in Eq.~\ref{equal:total_loss}, we use both long descriptions to enable VideoCLIP-XL to learn the semantics of long texts, and short descriptions to maintain the original short text ability.
For videos in our VILD dataset that do not have paired short descriptions from the origin resource, 
we use Qwen1.5-72B-Chat to generate them based on long descriptions. The prompt we used is:

\textit{``The following is a detailed video description. Please extract its core content and summarize it into a really short sentence. Do not exceed 10 words.}

\textit{\{examples\}}

\textit{Description: \{long video description\}}

\textit{Output:}''

For fine-tuned setting of text-video retrieval on traditional benchmarks, we tune our pre-trained VideoCLIP-XL with the vanilla text-video contrastive learning loss on each training set of the evaluated benchmarks.
During both training and testing, we sample 12 frames. Detailed hyper-parameters are the same as ViCLIP~\cite{viclip}. 
While in the zero-shot setting, along with the evaluations for Shot2Story and LVDR, we sample only 8 frames.

For the image CLIP models such as Long-CLIP, we calculate the similarity between the averaged image feature of frames and the text feature.

\begin{figure}[t]
\centering
\includegraphics[width=0.49\textwidth]{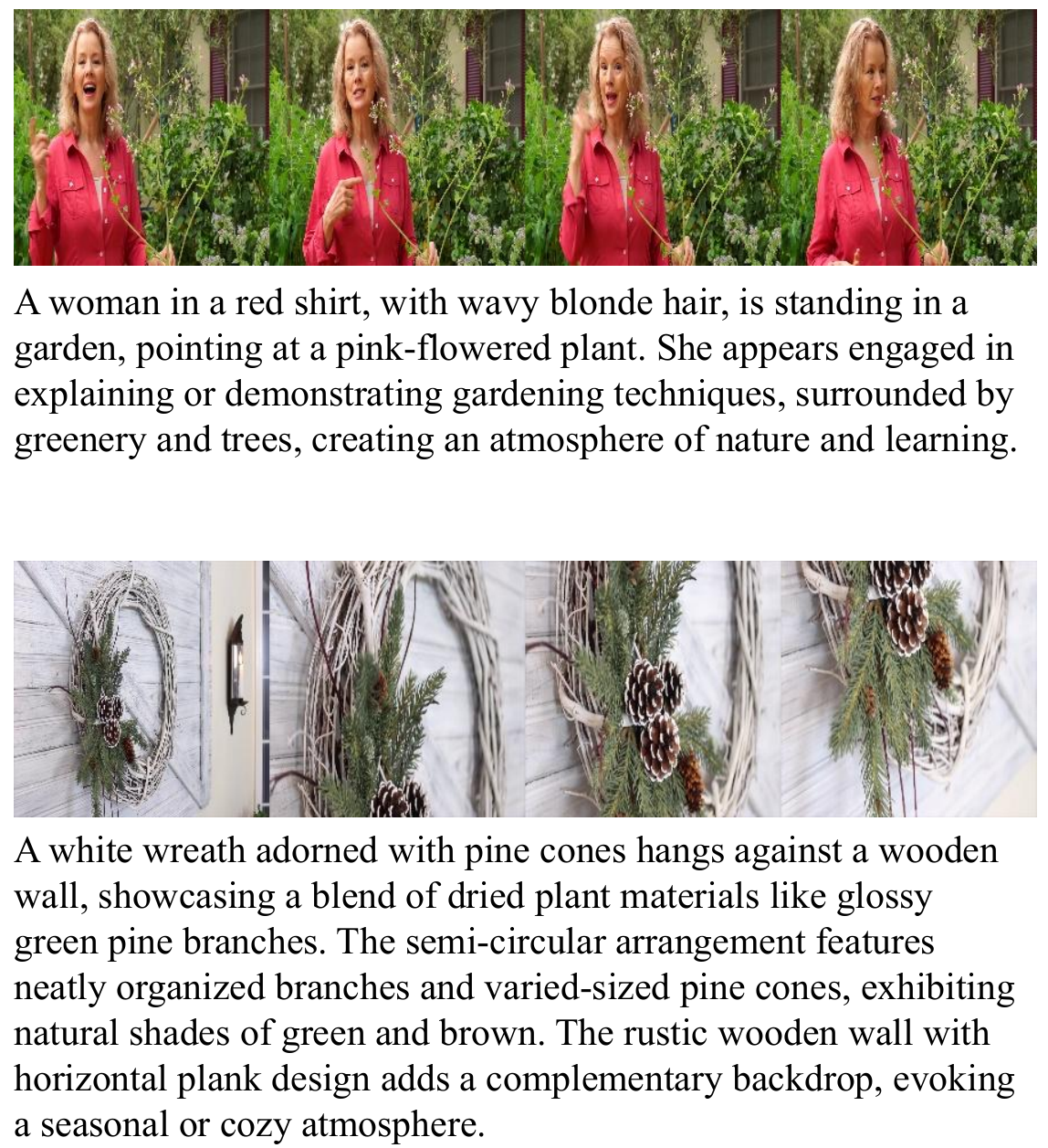} 
\caption{
Examples of synthetic long video captions in our VILD dataset.
}
\label{img:vld_example}
\end{figure}

\subsection{Performance Comparison with More Models}
As shown in Tab.~\ref{tab:appendix_more_result_compare},
we involve more recent powerful and large \textit{cross-encoder} models~\cite{umt,wang2024internvideo2} for comprehensive comparisons.
\textit{Cross-encoder} models, especially large multi-modal models (LMMs), typically add additional Transformer layers to model the deep interaction between vision and text representations.  
The model can generally boost the retrieval performance, while resulting in an unbearably slow retrieval speed when applied to the entire image/video collection since the cross-modal costs are required for each image/video sample whenever a new text query is given.
In contrast, our VideoCLIP-XL which has the \textit{dual-encoder}
structure has obviously fewer parameters and retrieval time cost.
\textit{Dual-encoder} encodes the visual and textual inputs in a wholly decoupled manner.
The vision representation is allowed to be pre-computed and re-used independent of the text queries.
Such approaches can utilize fast approximate nearest neighbor (ANN) search \cite{muja2009fast,jegou2010product,johnson2019billion} at runtime to ensure high efficiency.
For example, 
VideoCLIP-XL generally surpasses UMT-L~\cite{umt} on zero-shot text-video retrieval and has $\sim$4.14$\times$ faster retrieval speed on MSR-VTT without any bells and whistles, which can also indicate the effectiveness of our pre-training stage.
It is also $\sim$8.69$\times$ faster than InternVideo2$_{s2}$-1B.
For fine-tuning, large \textit{cross-encoder} models naturally surpass \textit{dual-encoder} models owing to the cross-modal feature interaction. Yet, these models still suffer from the low inference speed issue, and hence can hardly be deployed in real-time applications.
\begin{table*}[t]
\centering
\resizebox{\textwidth}{!}{%
\begin{small}
\begin{tabular}{@{}lcccccccccccc@{}}
\toprule
\multirow{2}{*}{Method} & \multicolumn{2}{c}{MSR-VTT} & \multicolumn{2}{c}{LSMDC} & \multicolumn{2}{c}{DiDeMo} & \multicolumn{2}{c}{MSVD} & \multicolumn{2}{c}{ActivityNet}  & \multicolumn{2}{c}{Avg.}\\
\cmidrule(l){2-13}
& T2V  & V2T  & T2V  & V2T  & T2V  & V2T  & T2V  & V2T  & T2V  & V2T   & T2V  & V2T  \\ 
\midrule
\textit{Dual-Encoder} & & & & & & & & & & & & \\
\midrule
CLIP~\cite{clip}      & 30.4 & 24.2 & 13.9 & 11.9 & 12.7 & 18.7 & 40.5 & 57.2 & 9.1 & 13.2 & 21.3 & 25.0 \\
VideoCLIP~\cite{videoclip} & 10.4  & - & -& -& 16.6 & -& -& -& -& -& -& - \\
CLIP4Clip~\cite{clip4clip} & 32.0 & - & 15.1 & - & - & - & 38.5 & - & - & -& -& -  \\
ViCLIP~\cite{viclip}      & 42.4 & 41.3 & 20.1 & 16.9 & 38.7 & 39.1 & 49.1 & 75.1 & 32.1 & 31.4 & 36.5 & 40.8  \\
\midrule
\tabincell{l}{\textbf{VideoCLIP-XL (Ours)}\\ \textbf{[}\textbf{V:304M/T:124M/C:0M/47.53s}\textbf{]}} & \textbf{50.1}  & \textbf{49.9}     & \textbf{22.8}  & \textbf{24.6}  & \textbf{47.7}  & \textbf{47.9}  & \textbf{51.9}  & \textbf{76.7}  & \textbf{46.4$^\ddag$}  & \textbf{48.1$^\ddag$} &\textbf{43.8} &\textbf{49.5}  \\
\midrule \rowcolor{gray!40}
\textit{Cross-Encoder} & & & & & & & & & & & & \\
\midrule\rowcolor{gray!40}
\rowcolor{gray!40}
\tabincell{l}{UMT-L~\cite{umt}\\ \textbf{[}\textbf{V:304M/T:271M/C:84M/196.68s}\textbf{]}}    & 40.7 & 37.1 & 24.9 & 21.9 & 48.6 & 49.9 & 49.0 & 74.5 & 41.9 & 39.4 & 41.0 & 44.6 \\
\rowcolor{gray!40}
\tabincell{l}{InternVideo2$_{s2}$-1B~\cite{wang2024internvideo2}\\\textbf{[}\textbf{V:1049M/T:271M/C:88M/413.09s}\textbf{]}}  & 51.9 & 50.9&  32.0&  27.3 & 57.0 & 54.3 & 58.1&  83.3&  60.4 & 54.8 &51.9 & 54.1\\
\rowcolor{gray!40}
\tabincell{l}{InternVideo2$_{s2}$-6B~\cite{wang2024internvideo2}\\\textbf{[}\textbf{NOT publicly available}\textbf{]}} &55.9& 53.7 &33.8 &30.1& 57.9 &57.1 &59.3& 83.1 &63.2 &56.5 & 54.0 & 56.1 \\
\bottomrule
\end{tabular}%
\end{small}
}
\\(a)\\
\bigskip

\resizebox{\textwidth}{!}{%
\begin{small}
\begin{tabular}{@{}lcccccccccccc@{}}
\toprule
\multirow{2}{*}{Method} & \multicolumn{2}{c}{MSR-VTT} & \multicolumn{2}{c}{LSMDC} & \multicolumn{2}{c}{DiDeMo} & \multicolumn{2}{c}{MSVD} & \multicolumn{2}{c}{ActivityNet}  & \multicolumn{2}{c}{Avg.}\\
\cmidrule(l){2-13}
& T2V  & V2T  & T2V  & V2T  & T2V  & V2T  & T2V  & V2T  & T2V  & V2T  & T2V  & V2T   \\ 
\midrule
\textit{Dual-Encoder} & & & & & & & & & & & & \\
\midrule
CLIP~\cite{clip}      & 38.2 & 38.7 & 22.5 & 22.6 & 32.2 & 33.9 & 52.3 & 69.9 & 26.1 & 26.9 & 34.3 & 38.4 \\
VideoCLIP~\cite{videoclip} & 30.9  & -& -& -& -& -& -& -& -& - & -& - \\
CLIP4Clip~\cite{clip4clip} & 45.6 & 45.9 & 24.3 & 23.8 & 43.0 & 43.6 & 45.2 & 48.4 & 40.3 & 41.6 & 39.7& 40.7 \\
ViCLIP~\cite{viclip}      & 52.5 & 51.8 & 33.0 & 32.5 & 49.4 & 50.2 & 53.1 & 79.0 & 49.8 & 48.1 & 47.6 & 52.3 \\
\midrule
\textbf{VideoCLIP-XL (Ours)}
& \textbf{57.0} & \textbf{56.6}  & \textbf{34.2}  & \textbf{32.6}  & \textbf{62.3}  & \textbf{62.7}  & \textbf{55.6}  & \textbf{81.4}  & \textbf{58.4$^\ddag$}  & \textbf{59.2$^\ddag$} & \textbf{53.5}  & \textbf{58.5}   \\
\midrule \rowcolor{gray!40}
\textit{Cross-Encoder} & & & & & & & & & & & & \\
\midrule\rowcolor{gray!40}
\rowcolor{gray!40}
UMT-L~\cite{umt}
&58.8& 58.6& 43.0 &41.4 &70.4 &65.7 &58.2 &82.4 &66.8& 64.4 & 59.4 & 62.5  \\
\rowcolor{gray!40}
InternVideo2$_{s2}$-6B~\cite{wang2024internvideo2}
& 62.8 & 60.2&  46.4 & 46.7 & 74.2 & 71.9 & 61.4 & 85.2&  74.1 & 69.7 & 63.8  & 66.7  \\
\bottomrule
\end{tabular}%
\end{small}
}
\\(b)
\caption{R@1 scores of (a) zero-shot and (b) fine-tuned text-video retrieval on MSR-VTT~\cite{xu2016msr}, LSMDC~\cite{lsmdc}, DiDeMo~\cite{didemo}, MSVD~\cite{msvd}, and ActivityNet~\cite{caba2015activitynet}.
$^\ddag$Due to the high overlap between the videos in ActivityNet and VideoInstruct100K~\cite{Maaz2023VideoChatGPT}, the latter is excluded from the pre-training data of our model tested on the former.
$\textbf{[}\textbf{V:304M/T:124M/C:0M/47.53s}\textbf{]}$ indicates that the vision encoder has 304M parameters, the text encoder has 124M parameters, the cross-encoder has 0M parameters, and this model needs 47.53s for text-video retrieval on MSR-VTT test set (1000 text-video pairs, tested on a single A100-80G GPU).
The same goes for others.
}
\label{tab:appendix_more_result_compare}
\end{table*}

\subsection{More Qualitative Results}

\begin{figure*}[t]
\centering
\includegraphics[width=0.99\textwidth]{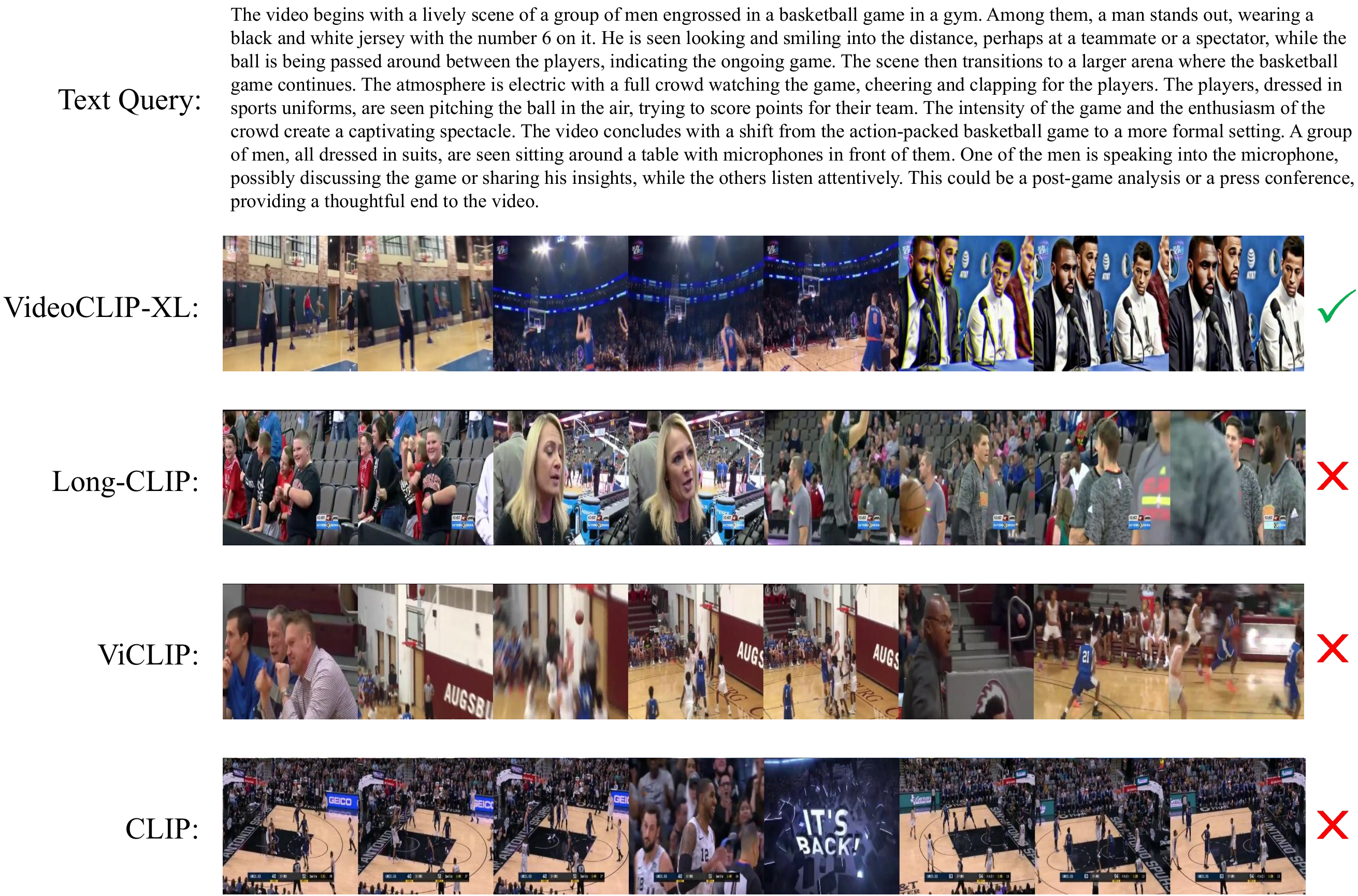}
\\(a)\\
\bigskip

\includegraphics[width=0.99\textwidth]{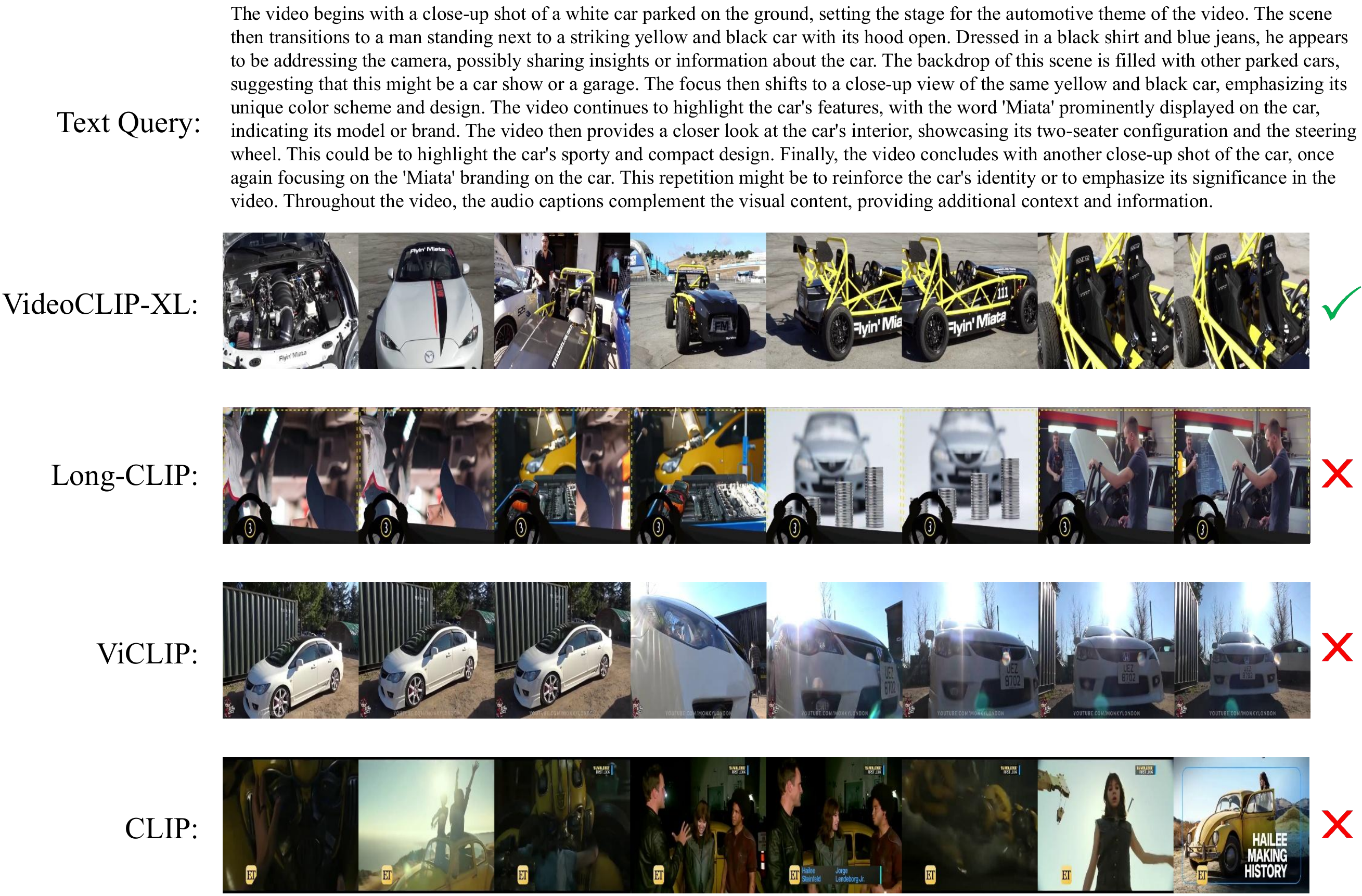} 
\\(b)\\
\caption{
Qualitative examples of text-to-video retrieval on the Shot2Story benchmark.
}
\label{img:qualitative_shot2story_example}
\end{figure*}

\label{qualitative_result_detail}
We give some examples of our synthetic long video descriptions acquired by Fig.~\ref{image-vld}(d)
in Fig.~\ref{img:vld_example}.
And qualitative examples of text-to-video retrieval results on the Shot2Story benchmark are shown in Fig.~\ref{img:qualitative_shot2story_example}. 
We can find that compared to competitors, our VideoCLIP-XL can achieve more accurate and matching video retrieval results.
% based on text queries.
\end{document}